\documentclass[journal]{IEEEtran} 
\pdfoutput=1

\usepackage[mathlines]{lineno}

\usepackage[T1]{fontenc}
\usepackage[utf8]{inputenc}
\usepackage{answers}
\usepackage{setspace}
\usepackage{graphicx}
\usepackage{multicol}
\usepackage{color}
\usepackage[table]{xcolor}
\usepackage{mathrsfs}
\usepackage[margin=1in]{geometry}
\usepackage{amsmath,amssymb}
\usepackage{tikz}
\usepackage{caption,subcaption}
\usepackage{mathtools}
\usepackage{listings}             
\usepackage{diagbox}

\newcommand\copyrighttext{%
\footnotesize Submitted to IEEE Journal of Oceanic Engineering (IEEE-JOE), under review as of June 2022. \copyright 2022 IEEE. Personal use of this material is permitted. Permission from IEEE must be obtained for all other uses, in any current or future media, including reprinting/republishing this material for advertising or promotional purposes, creating new collective works, for resale or redistribution to servers or lists, or reuse of any copyrighted component of this work in other works.}
\newcommand\copyrightnotice{%
\begin{tikzpicture}[remember picture,overlay]
\node[anchor=south,yshift=7pt] at (current page.south) {\fbox{\parbox{\dimexpr\textwidth-\fboxsep-\fboxrule\relax}{\copyrighttext}}};
\end{tikzpicture}%
}

\begin{document}
\title{Neural Shape-from-Shading for Survey-Scale Self-Consistent Bathymetry from Sidescan}
\author{\IEEEauthorblockN{Nils Bore \& John Folkesson}\\
\IEEEauthorblockA{Robotics, Perception and Learning Lab\\
Royal Institute of Technology (KTH)\\
Stockholm, SE-100 44, Sweden\\
Email: \{nbore, johnf\}@kth.se}}

\maketitle
\copyrightnotice
\vspace{-10pt} 

\begin{abstract}
Sidescan sonar is a small and cost-effective sensing solution
that can be easily mounted on most vessels. Historically, it
has been used to produce high-definition images that experts
may use to identify targets on the seafloor or in the water column.
While solutions have been proposed to produce bathymetry solely from sidescan, or in conjunction with multibeam, they
have had limited impact. This is partly a result of mostly being  limited to single sidescan lines. 
In this paper, we propose a modern, salable solution to create
high quality survey-scale bathymetry from many sidescan lines.
By incorporating multiple observations of the same place,
results can be improved as the estimates reinforce each other.
Our method is based on \textit{sinusoidal representation networks},
a recent advance in neural representation learning. We demonstrate the
scalability of the approach by producing bathymetry from a large sidescan survey.
The resulting quality is demonstrated by comparing to data collected with a high-precision
multibeam sensor.
\end{abstract}

\section{Introduction}
\label{introduction}

Maritime area surveys are usually carried out using acoustic sensors.
Because of the high attenuation of
electromagnetic signals underwater, acoustic waves propagate significantly
further at the same energy levels and therefore allow for longer range sensing.
Among sonar types, sidescan and multibeam are most commonly used to
survey larger seafloor areas. These two sensors exhibit some complementary properties,
which leads to them often being used in conjunction.
While multibeam returns three-dimensional positions, allowing for the creation of georeferenced
height maps, sidescan only measures return backscatter intensities. Furthermore, sidescans
are generally high-resolution compared to multibeam and have a wider
swath range, thus surveying a larger area. The fine resolution of the sidescan
can often be exploited by skilled sonar operators in order to detect objects on the seafloor.

While sidescan cannot be used to directly reconstruct geometry, it is clear
that the signal does contain information about the slope of the seafloor.
Indeed, a large portion of the signal can be well approximated by a Lambertian illumination
model \cite{bell1999sidescan}. 
Because of this, there has been several attempts at bathymetric
reconstruction from sidescan images, most notably using
shape-from-shading techniques \cite{coiras2005expectation}.
All such methods inevitably face the problem of the sonar geometry.
Sidescan emits a narrow beam along the direction of travel. However, the width
orthogonal to this direction is very wide, resulting in the mentioned wide swath range.
For each reflected intensity, the sensor measures only the two-way travel time,
resulting in the angle of the recorded reflection being unknown.
Typically, methods solve this by integrating the surface from a known or estimated
point close to the vehicle in order to reconstruct the seafloor.
However, even if the sonar is looking at a simple geometry such as a flat seafloor,
it is difficult to model the sonar to such an extent that you can extract exact surface slopes.
This results in the accuracy of the height estimate drifting the further away from
the sensor you get.

\begin{figure*}[t]
 \begin{center}
 \input{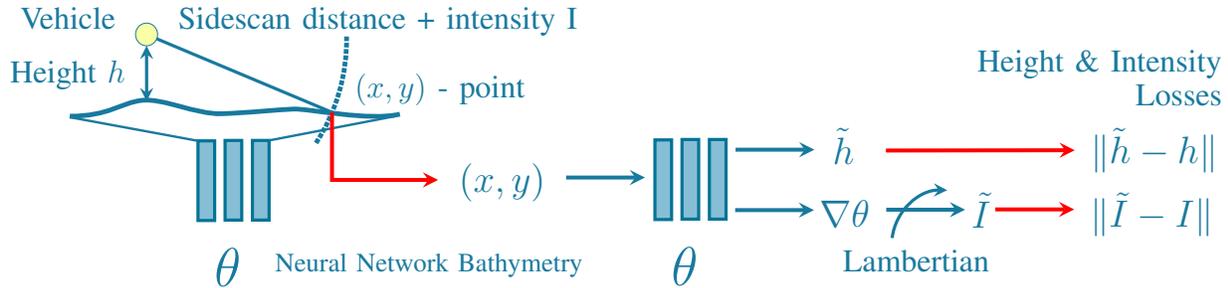}
 \caption{Our method learns a bathymetry height map $\theta$ from sidescan intensities $I$
          and altimeter heights $h$. $\theta$ is a
 		  differentiable mapping from 2D $x, y$ positions to seafloor heights $h$. When training,
 		  sidescan intensities are mapped to $x, y$ points by finding intersections of isotemporal
 		  curves with the analytic representation $\theta$. The method minimizes the distance to sparsely
 		  measured altimeter points $h$. From the gradient $\nabla \theta$, it also computes Lambertian
 		  intensities which are compared with the ground truth sidescan intensities. These aid the
 		  network in interpolating in between measured altimeter points.}
  \label{fig:method_overview}
 \end{center}
\end{figure*}

\begin{figure*}[thpb]
	\centering
  	\includegraphics[clip, trim=2.5cm 0.5cm 2.5cm 1.5cm, width=0.99\linewidth]{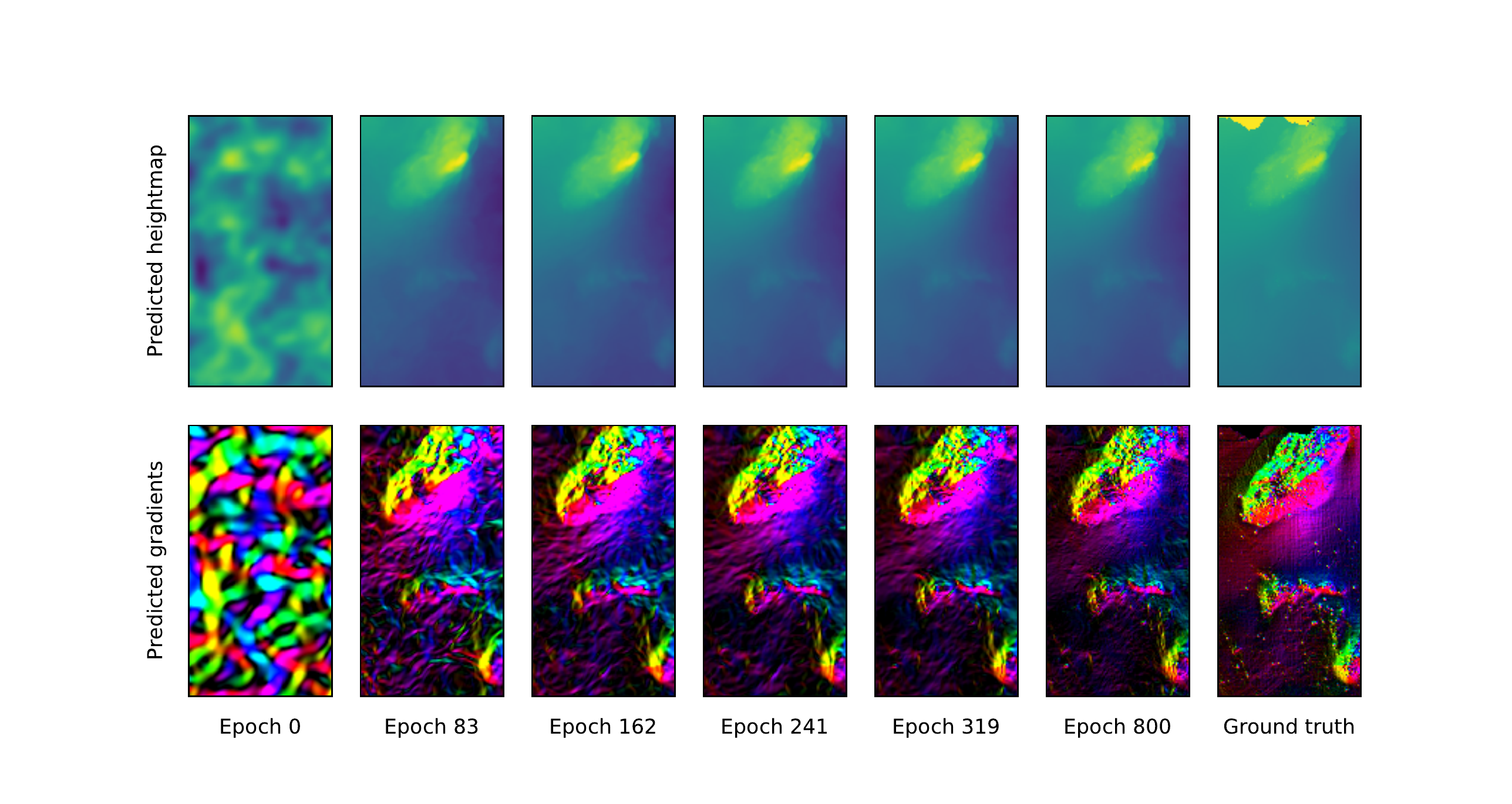}
 	\caption{Illustration of the progression of the training of our model, with height maps
 			 in the top row and corresponding gradients below.
 			 The model is initialized with random parameters. As training progresses,
 	         it quickly captures the large hill at the top of the image, as well as two smaller ones below. 
 	         Around epoch 300, the model also starts to identify smaller rocks.
 	         At the far right, the corresponding ground truth bathymetry and gradients are presented.}
 	\label{fig:training_imgs}
\end{figure*}

Much like traditional shape-from-shading using real-world camera images,
bathymetry reconstruction using single sidescan lines is an underconstrained problem.
There is another line of research, often referred to as \textit{inverse rendering},
that attempts to reconstruct a three-dimensional scene geometry using multiple calibrated camera images.
With more images, it becomes tractable to pose this as an optimization
problem, wherein a geometry is estimated that best reproduces all of the captured images.
In recent years, that field has moved towards \textit{differentiable rendering}, and more
recently, neural networks as implicit representations of the three-dimensional geometry.
This latter field of research is often referred to as \textit{representation learning},
and neural models such as \cite{mildenhall2020nerf}\cite{sitzmann2019siren} produce state-of-the-art results.
The differentiable nature of these models also present some tantalizing possibilities with
regard to sidescan sonar. In particular, neural models such as \cite{sitzmann2019siren} allow
constraining the slope, or gradient, to match the one inferred from the sidescan intensity.

In order to reduce the impact of the estimate drifting when
integrating the gradients, we propose also including some depth measurements.
These may come from an altimeter echosounder or from bottom-tracking
the sidescan signal itself \cite{yan2020real}. The additional information
greatly aids in the reconstruction as it allows the surface to be ``pinned down''
at certain points.
In addition, we argue that it is critical that bathymetry is predicted
jointly from multiple sidescan lines at the same time.
By phrasing it as
a global optimization problem, and by employing recent computer vision
advances, we can estimate the solution in a practical fashion.
In doing so, the sidescan lines aid each other by reducing the uncertainty
in seafloor height, thus letting the method focus also on reconstructing
smaller scale detail such as rocks with high precision.

A method overview is illustrated in Figure \ref{fig:method_overview}.
Our method receives as input sidescan intensities and ranges, as well
as sparse depth measurements from an altimeter sensor.
The bathymetry is represented as a neural network that maps
two-dimensional positions to predicted heights. Since the representation is
by its nature differentiable, it can also return surface gradients.
With a simple Lambertian illumination model it can then generate
predicted intensities that can be compared with the sidescan intensities.
Indeed, since the intensities only put constraints on the gradient values,
one can think of the model as optimizing a system of differential
equations. The altimeter measurements provide the boundary conditions
by allowing us to fix the solutions at some sparse points.
Figure \ref{fig:training_imgs} illustrates the optimization
procedure, starting from some random initial values.

We present the following contributions:
\begin{enumerate}
\item A novel, differentiable bathymetry representation
\item A framework for combining multiple sidescan lines into a self-consistent bathymetric model
\item Experiments that validate the scalability of the approach when applied to a large sidescan survey
\end{enumerate}

\subsection{Related work}
\label{related_work}

Before discussing the problem of bathymetry estimation from sidescan,
it will be instructive to first consider the inverse problem of simulating sidescan.
As it is often simpler to reconstruct sidescan from bathymetry, this is usually the way in
which the former problem is phrased, with one notable exception being data-driven
models \cite{xie2019inferring}.
Model-driven approaches
\cite{williams1998bistatic} 
\cite{bell1997simulation}
\cite{dura2004reconstruction} 
\cite{coiras2009gpu}
\cite{pailhas2009realtime}
take as input seafloor geometry,
possibly together with some relevant acoustic properties.
They then ``simulate'' a sidescan by computing where the emitted beam could
intersect the seafloor and reflect back in an amount of time corresponding to the
temporal resolution of the sonar.
The geometry as well as potential seafloor properties
are used to reconstruct a signal. Most works assume some form of diffuse reflection
model \cite{folkesson2020lamberts}, with varying degrees of higher-order corrections added on top
\cite{williams1998bistatic}\cite{bell1997simulation}.
Again, there are examples of different, model-free approaches,
including recent data driven models \cite{bore2021modeling}.

Many of the methods for solving the opposite, inverse, problem can be characterized
as \textit{shape-from-shading}. This technique was pioneered by
Horn \cite{horn1975obtaining} to estimate geometry of diffusely reflecting
materials from single camera images. Woodham \cite{woodham1980photometric} proposed
a similar method with multiple calibrated cameras, alleviating reliance
on the model assumptions being perfectly fulfilled.
Langer \& Hebert \cite{langer1991building} were among the first to apply
similar techniques to sidescan imagery. Similar to many subsequent methods,
their method starts by removing any points corresponding to shadows, since
they cannot be explained by the reflection model. In addition, they factor
the reflection model into diffuse and specular components, similar to
Jackson's sonar model \cite{williams1998bistatic}.
Dura et al. \cite{dura2004reconstruction} compared Langer \& Hebert's
method to calculating geometry through the inverse of Bell's method \cite{bell1997simulation}.
In essence, since the latter is a linear operation on the Fourier transform of the
sidescan signal, the inverse can be easily computed.
They conclude that the two approaches are suitable in different scenarios,
with their linear reflectance map approximation being more robust to noisy signals.
Johnson \& Hebert \cite{johnson1996seafloor} 
provide a good overview of these early techniques for solving the
inverse problem. They propose an algorithm that augments the sidescan
bathymetry with sparse direct bathymetric measurements.
Specifically, they initialize the bathymetry estimate with the
sparse measurements and perform a global optimization over imagery
from one full survey line.

In a series of notable papers, Coiras et al.
\cite{coiras2005expectation}
\cite{coiras2007multiresolution}
proposed a method to estimate not only the seafloor geometry,
but also the surface albedo as well as the beam-pattern.
Their method works solely from sidescan backscatter, and uses
an expectation-maximization approach to alternately estimate
these properties and then reconstruct the sidescan signal.
To improve convergence, they gradually increase the resolution
of the predicted bathymetry \cite{coiras2007multiresolution}, with
a lower resolution version used to initialize the next step.
In particular, their results demonstrate the ability of the method
to estimate all three properties realistically without
significant overfitting to individual data points.

In one of the earliest systems for bathymetry-from-sidescan, 
Cuschier \& Hebert
\cite{cuschieri1990three} 
proposed an approach different from shape-from-shading.
Their system identified shadows within the sidescan
pings and used trigonometry to compute the height of the protrusions causing them.
Similar to many subsequent shape-from-shading approaches, they then integrate
the contributions from the individual slices to form the full seafloor profile.
Notably, Bikonis et al.
\cite{bikonis2008computer} 
combined shape-from-shading with this sort of shadow information in order
to extract improved depth estimates.
In a conceptually similar but more elaborate approach,
Woock \& Beyerer \cite{woock2014seafloor} 
proposed deconstructing the sidescan signal into a ``library''
of signal templates, each with an associated bathymetry profile.
By combining the contributing profiles, a full depth profile can then
be reconstructed. While more complete than the simple shadow detection
approach of \cite{cuschieri1990three}, it faces similar challenges in
reconstructing general types of seafloors.
Recently, Jones \& Traykovski \cite{jones2018method} 
employed shadow geometry to reconstruct the surface surrounding
a rotating sidescan mounted close to the seabed. They perform thorough
evaluations of the efficiency of their method, with quantitative comparisons
to multibeam bathymetry.

While many early works \cite{cuschieri1990three}\cite{langer1991building}
focus on estimating geometry mostly from sidescan, they generally note that
the problem is ill-defined without some additional assumptions or boundary conditions.
Li \& Pai
\cite{rongxing1991improvement} 
suggest a different route by augmenting multibeam data with sidescan. 
Their method works on a grid-based terrain model, with each grid cell forming
the boundary of a shape-from-shading problem in between the multibeam points.
With a typical high-resolution sidescan, the resolution of the terrain model
can thus be enhanced.
Johnson \& Hebert's approach \cite{johnson1996seafloor} 
is more general in that it only relies on depth measurements wherever available.
They optimize a local surface in sidescan coordinates. The optimized surface is then
used to associate the sidescan bins with grid cells in a global coordinate system.
In the final step, the grid representation is optimized using a non-linear
minimization technique similar to Horn's \cite{horn1986robot}.
The sparse depths are used to initialize the height map, but only
smoothness regularization is used as a constraint.
Johnson \& Hebert's approach \cite{johnson1996seafloor} is notable
in that it is \textit{in principle} able to deal with observations originating
from multiple sidescan lines. Most other methods for
constructing aggregate bathymetric maps  
\cite{woock2011deep}
\cite{zhao2018reconstructing} 
have not dealt explicitly with this problem. Instead, they have proposed
forming submaps for different lines \cite{woock2011deep} or simply
averaged the individual geometries \cite{zhao2018reconstructing}.

Looking more generally at sonar geometry, several methods for reconstructing three-dimensional geometry
from wide-aperture sonar have been proposed. As 2D imaging sonars cover a large
volume with significant overlap in-between subsequent measurements, it is viable to use
feature-based structure-from-motion \cite{huang2015towards}\cite{brahim20113d}
to estimate sensor motion as well as sparse geometry.
To form dense object geometry, space-carving methods \cite{aykin2013forward}\cite{aykin2017three}
may be used. However, they work poorly for non-convex geometries,
including most seafloor geometries of interest.
More recently, works such as  
\cite{westman2019wide}
\cite{westman2020volumetric}
\cite{guerneve2018three} 
applied more detailed backscatter models, resulting in higher-quality
reconstructions \cite{westman2019wide}\cite{guerneve2018three}.
Guerneve et al. \cite{guerneve2018three} suggested a method for resolving
high-resolution surface structures from wide-aperture sonar by means of
a spatially varying deconvolution approach. The method is fast but constrained
to motion along the wide beam angle of the sonar (typically corresponding to the z-axis).
To infer geometry from unconstrained motion,
Westman et al. \cite{westman2020volumetric} proposed inferring
an acoustic ``albedo'' field for the full surveyed volume.
by means of a \textit{non-line-of-sight} (NLOS) reconstruction.
Since the field is a scalar field that is uniform in all directions,
it does not take geometric surface scattering into account.
A synthetic aperture tomography approach was also presented
in \cite{marston2016volumetric}. 
The method extracts high-definition models of relatively small objects
from data collected by an autonomous underwater vehicle equipped with a \textit{synthetic aperture sonar} (SAS) sensor.

One important aspect that differentiates the surveyed methods is how they
solve the optimization problem resulting from trying to reconstruct the backscatter.
Dura et al. \cite{dura2004reconstruction} characterize the early methods as
either global optimization, propagation out from a known height or as linear approaches.
The vast majority of later approaches \cite{johnson1996seafloor}
\cite{coiras2007multiresolution}\cite{bikonis2008computer}
\cite{zhao2018reconstructing}\cite{westman2020volumetric}
can be characterized as global optimization, meaning that they
optimize the full ping or image jointly. The difference is often in the
solution to the optimization problem, with methods using for example
general non-linear solvers \cite{johnson1996seafloor}, 
sparse solvers \cite{westman2020volumetric}
or expectation-optimization with gradient descent \cite{coiras2007multiresolution}.
An advantage of global optimization is that priors such
as smoothness on the produced bathymetry can be enforced,
thus increasing robustness to sensor noise or modeling errors.
So far, no method has been proposed that utilizes the GPU-accelerated optimization
of modern deep learning libraries.

Our method presents a scalable solution to the shape-from-shading
problem from multiple sidescan lines.
The mathematical model is similar to that of Coiras et al. \cite{coiras2007multiresolution}.
However, we employ a neural network as the bathymetric representation,
similar to recently proposed representation learning systems \cite{sitzmann2019siren}.
In contrast to \cite{coiras2007multiresolution}, our method combines many
sidescan lines into a self-consistent, georeferenced bathymetry. By optimizing
all of the lines jointly using the GPU, their individual estimates can reinforce each other
to remove some of the inherent ambiguity of the inverse problem.
Our optimization scheme itself is somewhat similar to the reconstruction
from forward-looking sonar described by Westman et al. \cite{westman2020volumetric}.
An important advantage of our method is that our model incorporates the 
geometric backscattering through the incidence angle.
Johnson \& Hebert's approach \cite{johnson1996seafloor} is similar to
ours in that it in principle allows for several sidescan lines, and also
accommodates sparse depths. In contrast to that paper and all previous work,
we demonstrate our method on many sidescan lines, from a large area survey,
and jointly optimize the bathymetry to reconstruct all the backscatter data.
In addition, we provide an extensive quantitative evaluation where we compare
the produced bathymetric map to high-precision multibeam bathymetry.

\subsection{Problem statement}

The goal of the described method is to construct high-precision
bathymetry from many sidescan survey lines together with some sparse depth measurements.
We assume that the global pose of the sidescan and depth sensors are
known, either through GPS or some other means of localization, such as
\textit{simultaneous localization and mapping} (SLAM).

\section{Method}
\label{method}

We begin by defining our bathymetric representation $\theta$ as well as
the rather general loss terms that we use to optimize it.
Subsequently, we will describe how to simulate sidescan
observations given this bathymetric representation in a differentiable manner.

\subsection{Loss functions}

Our method aims to find a function $\theta: \mathbb{R}^2 \rightarrow \mathbb{R}$
that maps two-dimensional Euclidean world coordinates to the respective seafloor height.
As will become clear, it is crucial that $\theta$ is differentiable. While $\theta$
could in principle be any trainable, differentiable black box method, we will take it to be
an \textit{multi-layer perceptron} (MLP) variant known as a \textit{Sinusoidal Representation Network} (SIREN) \cite{sitzmann2019siren}. The method has the advantage of giving
high-quality derivatives, as well as not suffering from ``spectral bias''
\cite{tancik2020fourfeat}. 
The latter feature allows the network to also learn smaller, high-frequency, features of the seafloor.

\definecolor{lightgray}{gray}{0.9}
\begin{table}[htpb]
\begin{center}
\rowcolors{1}{}{lightgray}
\begin{tabular}{r|c}
  Inputs & Description \\
  \hline
  $t_i \in \mathbb{R}^3$ & sidescan origin at ping $i$ \\
  $\mathbf{R}_i \in SO(3)$ & sidescan rotation matrix at ping $i$ \\
  $I_{i, n} \in \mathbb{R}$ & intensity at ping $i$ and time bin $n$ \\
  $p^{gt}_j \in \mathbb{R}^3$ & measured seafloor height \\
  Outputs &  \\
  \hline
  $\theta: \mathbb{R}^2 \rightarrow \mathbb{R}$ & bathymetry height map \\
  $R: \mathbb{R}^2 \rightarrow \mathbb{R}_{+}$ & albedo map \\
  $\Phi: \mathbb{R} \rightarrow \mathbb{R}_{+}$ & beam-pattern \\
  $A_i$ & per-line gain \\
\end{tabular}
\end{center}
\caption{The main inputs and outputs of our method.}
\label{tab:variables}
\end{table}

Our method aims to reconstruct bathymetry from sparse altimeter readings, as well as dense
sidescan imagery.
In addition, for this study we rely on high-quality positioning,
see Table \ref{tab:variables} for all inputs to the method.
We will therefore have two parts to our loss function,
with $\alpha$ determining their relative importance,
\begin{linenomath*}
\begin{equation}
\label{eq:global_loss}
    \mathcal{L} =  \mathcal{L}_{\nabla} + \alpha \mathcal{L}_{H}.
\end{equation}
\end{linenomath*}

The $\mathcal{L}_{H}$ loss from the altimeter readings $p^{gt}$ simply aims to minimize the distance from
$\theta$ at coordinates $(p^{gt}_x, p^{gt}_y)$ to the measured height $p^{gt}_z§$.
If we take 
\begin{linenomath*}
\begin{equation}
    \Delta^\theta(p) = \theta(p_x, p_y) - p_z
\end{equation}
\end{linenomath*}
to be the signed vertical distance, the averaged loss from
one batch of altimeter heights is given by
\begin{linenomath*}
\begin{equation}
    \mathcal{L}_{H} = \frac{1}{\left| \left\{ p^{gt}_j \right\} \right| }\sum_{j} \| \Delta^\theta(p^{gt}_j) \|.
\end{equation}
\end{linenomath*}

The second part, $\mathcal{L}_{\nabla}$, aims to reconstruct the ground truth intensity $I_{i, n}$ from the sidescan
for each sidescan ping $i$ and each distance bin $n$, 
\begin{linenomath*}
\begin{equation}
    \mathcal{L}_{\nabla} = \frac{1}{\left| \left\{ I_{i, n} \right\} \right|}\sum_{{I_{i, n}}} \left\Vert \: \tilde I_{i, n} - I_{i, n}  \: \right\Vert.
\end{equation}
\end{linenomath*}
The remainder of the paper mainly deals with the difficult problem of estimating
the intensities $\tilde I_{i, n}$ from the bathymetry $\theta$.

\subsection{Computing seafloor intersections}

In a sidescan, the beam width in the longitudinal direction is small
enough to mostly be neglected. Spatially, this leaves the reflected volume around a thin
arch at a fixed time away from the vehicle. We will refer to such archs as
\textit{isotemporal curves}.
If we assume an isovelocity \textit{sound velocity profile} (SVP), the isotemporal curve
also corresponds to a fixed distance, parameterized by an angle $\phi$,
\begin{linenomath*}
\begin{equation}
    p_{i, n}(\phi) = t_i + d_n R_i \left[0, \sin(\phi), -\cos(\phi)\right]^T.
\end{equation}
\end{linenomath*}
While, for simplicity, we assume an isovelocity
SVP in this paper, $p_{i, n}$ could in
principle be any differentiable function whose path approximates
an isotemporal curve as defined by an SVP.

Additionally, we will assume that the function $\Delta^\theta(p_{i, n}(\phi_k))$
only has one zero-crossing within the interval $\left[\phi_{min}, \phi_{max}\right]$.
If its derivative is strictly positive around that crossing,
we may find it using gradient descent optimization. Such a scheme iteratively
updates the angular position $\phi$ from a starting value $\phi^0 = \frac{1}{2}\left( \phi_{min} + \phi_{max} \right)$, where $\phi_{min}$ and $\phi_{max}$
define the boundaries of the beam,
\begin{linenomath*}
\begin{equation}
\label{eq:gradient_descent}
\phi^{k+1} = \phi^k - \frac{\lambda}{d_{i, n}} \frac{d}{d\phi} \left( \Delta^\theta(p_{i, n}(\phi^k)) \right)^2.
\end{equation}
\end{linenomath*}
The $\lambda$ parameter describes the update step size, while division by the distance
$d_{i, n}$ ensures that the step is comparable across the different curves.
We perform a fixed number of gradient descent steps, resulting in the angle
$\phi_{i, n}^*$ at the seafloor intersection.
If $\phi_{i, n}^* \not\in \left[\phi_{min}, \phi_{max} \right]$, we exclude
it from the loss computation, since no zero-crossing was found within the beam.

\subsection{Lambertian scattering model}

Subsequently, we will focus on ways of inferring the model intensity
$\tilde I_{i, n}$ from the height map model $\phi$ assuming a Lambertian scattering model.
To compute a Lambertian intensity, we first need to define the normal with respect to $\theta$
at a point $p \in \mathbb{R}^3$. Given the two gradient components $\nabla_x, \nabla_y$ it is defined by
\begin{linenomath*}
\begin{equation}
    N^\theta(p) = \left[ -\nabla_x \theta(p_x, p_y), -\nabla_y \theta(p_x, p_y), 1 \right]^T.
\end{equation}
\end{linenomath*}

In the general case, the incident ray of the isotemporal curve $p_{i, n}$ is given by the normal to
its path derivative at the point of intersection with the seafloor.
Assuming a port-facing sidescan, the rotation from the tangent to the normal is given by the $90^{\circ}$ rotation around the x-axis in the sensor frame, $R_i R_x(\frac{\pi}{2}) R_i^T$, giving
\begin{linenomath*}
\begin{equation}
    r_i(\phi) = R_i R_x(\frac{\pi}{2}) R_i^T \frac{d}{d \phi} p_{i, n}(\phi)
\end{equation}
\end{linenomath*}
and specifically, for an isovelocity SVP,
\begin{linenomath*}
\begin{align*}
    r_i(\phi) &= d_n R_i R_x(\frac{\pi}{2}) \frac{d}{d \phi} \left[0, \sin(\phi), -\cos(\phi)\right]^T \\
              &= d_n R_i R_x(\frac{\pi}{2}) \left[0, \cos(\phi), sin(\phi)\right]^T \\
              &= d_n R_i \left[0, -\sin(\phi), cos(\phi)\right]^T.
\end{align*}
\end{linenomath*}

Using $N^\theta$ and $r_i$, we can compute the cosine of the incidence angle, giving
the Lambertian scattering contribution $M_{i, n}^\theta$ from one point $p$.
More specifically, we use the $cos^2$ approximation,
as it is known to give better results at low gracing angles:
\begin{linenomath*}
\begin{equation}
    M_{i, n}^\theta(\phi^*) = \left( r_i(\phi^*)^T \hat N^\theta(p_{i, n}(\phi^*)) \right) ^2.
\end{equation}
\end{linenomath*}

\subsection{Beam-pattern, gain and albedo models}

In addition to the surface normals implicitly given by the Lambertian intensity,
we also estimate other important parameters of the sensor and its environment.
Our full model is inspired by that of Coiras et al. \cite{coiras2007multiresolution}.
In addition to the albedo $R$ of the seafloor and the beam-pattern $\Phi$, we also
estimate  a gain parameter $A_i$. Each of these parameters are initialized to $1$
when training, and are constrained to be
positive by passing them through an $exp$ function before feeding them into the model.
The complete intensity model is given by
\begin{linenomath*}
\begin{equation}
\label{eq:intensity}
\tilde I_{i, n} = K A_i M_{i, n}^\theta(\phi_{i, n}^*) \Phi(\phi_{i, n}^*)  R(p_{i, n}(\phi_{i, n}^*)).
\end{equation}
\end{linenomath*}
Within this context, $K$ is a normalization constant that is
estimated once for the whole data set prior to training.
While $K$ is there purely to help the optimization converge faster (see Section \ref{sec:data_preparation} for details),
the rest of the parameters are optimized during training together with $\theta$.
The gain parameter $A_i$ is estimated for each sidescan line, and allows for varying gain
across the data set. The functions $\Phi(\phi)$ and $R(p)$ are kernel densities
whose kernel weights $\Phi_\kappa, R_\kappa$ are estimated while training. The spread $\sigma$ of the fixed position kernels
are taken to by the range divided by the number of kernels, with $\gamma = \frac{1}{2 \sigma^2}$.
Thus, the beam-pattern is defined as
\begin{linenomath*}
\begin{equation}
\Phi(\phi) = \frac{1}{Z} \sum_{\kappa} \Phi_\kappa \exp(- \gamma_\Phi \left( \phi_\kappa - \phi \right)^2),
\end{equation}
\end{linenomath*}
while the 2-dimensional albedo function is given by
\begin{linenomath*}
\begin{equation}
R(p) = \frac{1}{Z} \sum_{\kappa} R_\kappa \exp(- \gamma_R \| x_\kappa - p_{xy} \|^2).
\end{equation}
\end{linenomath*}
Similar to \cite{johnson1996seafloor}, we use a rather coarse approximation of $R_{i, n}$,
with only 100 kernels spaced in a grid. This ensures that we do not fit variations appearing
only in few sidescan lines, which may be due either to geometry or other factors.

\subsection{Optimization}

The proposed representation $\theta$ as well as the kernel densities
$R, \Phi$ are fully differentiable. However, the gradient descent
procedure of Equation \ref{eq:gradient_descent} is not easily differentiable.
Instead, for every training batch, we compute the intensity $\tilde I_{i, n}$ in Equation \ref{eq:intensity} from the produced point $p_{i, n}(\phi^*)$,
without backpropagating through the gradient descent procedure.
As soon as training of the representation $\theta$ has converged to be roughly within
the true height range, this simplification should have a negligible impact.

With these details out of the way, the loss function of Equation \ref{eq:global_loss} can be optimized with a standard neural network optimizer;
in the experiments we use ADAM. Again, it is worth noting the similarities to a differential
equation with boundary conditions. The intensity implicitly defines the
gradients at some points while the altimeter readings provide the boundary
points. For the optimization to converge to a reasonable bathymetry, we
have observed that it requires at least some altimeter readings,
and preferably at the perimeter.

\section{Experiments}
\label{experiments}

\subsection{Data set}

Our method was evaluated on a data set that also contains high-precision bathymetry
that can be used as comparison. The sidescan data was collected using a surface
vessel equipped with \textit{RTK GPS}, ensuring high quality positioning.
Similar to most AUVs, the sidescan is hull
mounted, thus ensuring that its position is accurate as well.
Our system takes as input the raw sidescan files produced by the surveys.
To facilitate the evaluation of our sidescan modeling, we used the high-precision bathymetry
to simulate unambiguous altimeter readings.
Since they already agree with the compared bathymetry, this ensures that we evaluate only
the effectiveness of our sidescan modeling, thus removing any potential artifacts from
an actual altimeter sensor.

\definecolor{lightgray}{gray}{0.9}
\begin{table}[htpb]
\begin{center}
\rowcolors{1}{}{lightgray}
\begin{tabular}{r|c} 
  Property & Value \\ 
  \hline
  Bathymetry res. & $0.5m$ \\ 
  Sidescan type & Edgetech 4200MP \\ 
  Sidescan range & $0.035s \Rightarrow \sim 50m$ \\ 
  Sidescan freq. & $555 kHz$ \\ 
  Composition & $\sim 70\%$ Sed. rock, $30\%$ sand \\ 
  Mean altitude & $17m$ \\ 
  Survey area & $\sim 350m \times 300m$ \\ 
  Sidescan pings & $\sim 93000$ \\ 
\end{tabular}
\end{center}
\caption{Data set and survey area characteristics.}
\label{tab:datasets}
\end{table}

When comparing our height map estimate to the high-precision bathymetry, we consider
several metrics. Each of them is computed at the native $0.5m$ scale of the bathymetry.
In addition to the mean absolute height difference between the two, we also compare
the gradients. The reason for studying the gradients is that the
sidescan signal contains information that directly relates to the surface gradients. However,
when integrating these gradients to produce the ground truth height map, the estimate will
drift. By considering both absolute height and gradients, we can therefore evaluate
gradient quality as well as the drift due to integration. The multibeam
bathymetry gradient is calculated from finite differences. It is compared both by looking
at the cosine similarity to the predicted gradients, as well as the mean absolute gradient
magnitude difference.

\subsection{Data preparation}
\label{sec:data_preparation}

\begin{figure*}[htpb]
\centering
\begin{subfigure}{.49\textwidth}
  \centering
  \includegraphics[clip, trim=2.2cm 0.1cm 3.2cm 1.4cm, width=.99\linewidth]{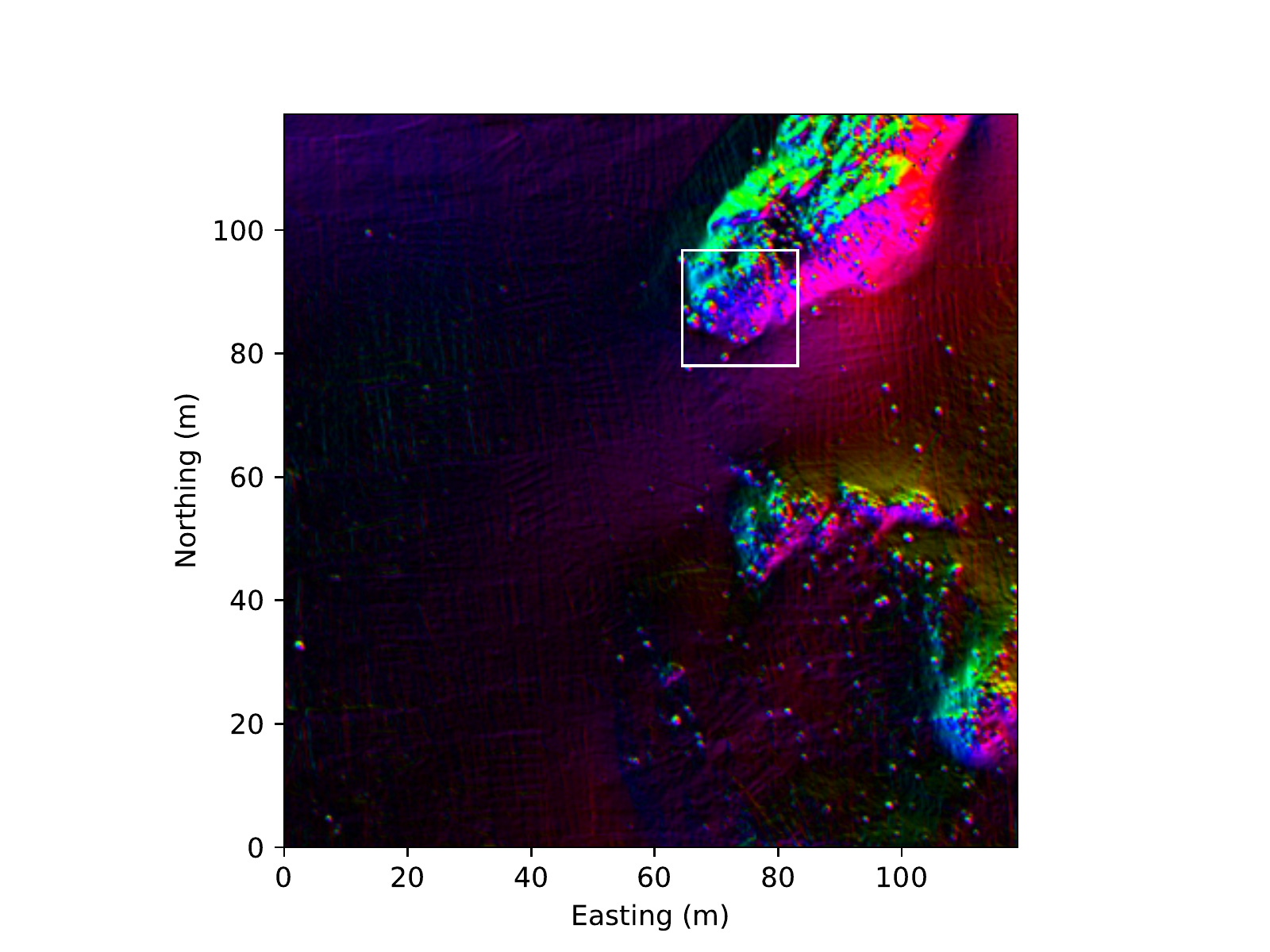}  
  \caption{Ground truth gradients.}
  \label{fig:densities_gt}
\end{subfigure}
\begin{subfigure}{.49\textwidth}
  \centering  
  \includegraphics[clip, trim=2.2cm 0.1cm 3.2cm 1.4cm, width=.99\linewidth]{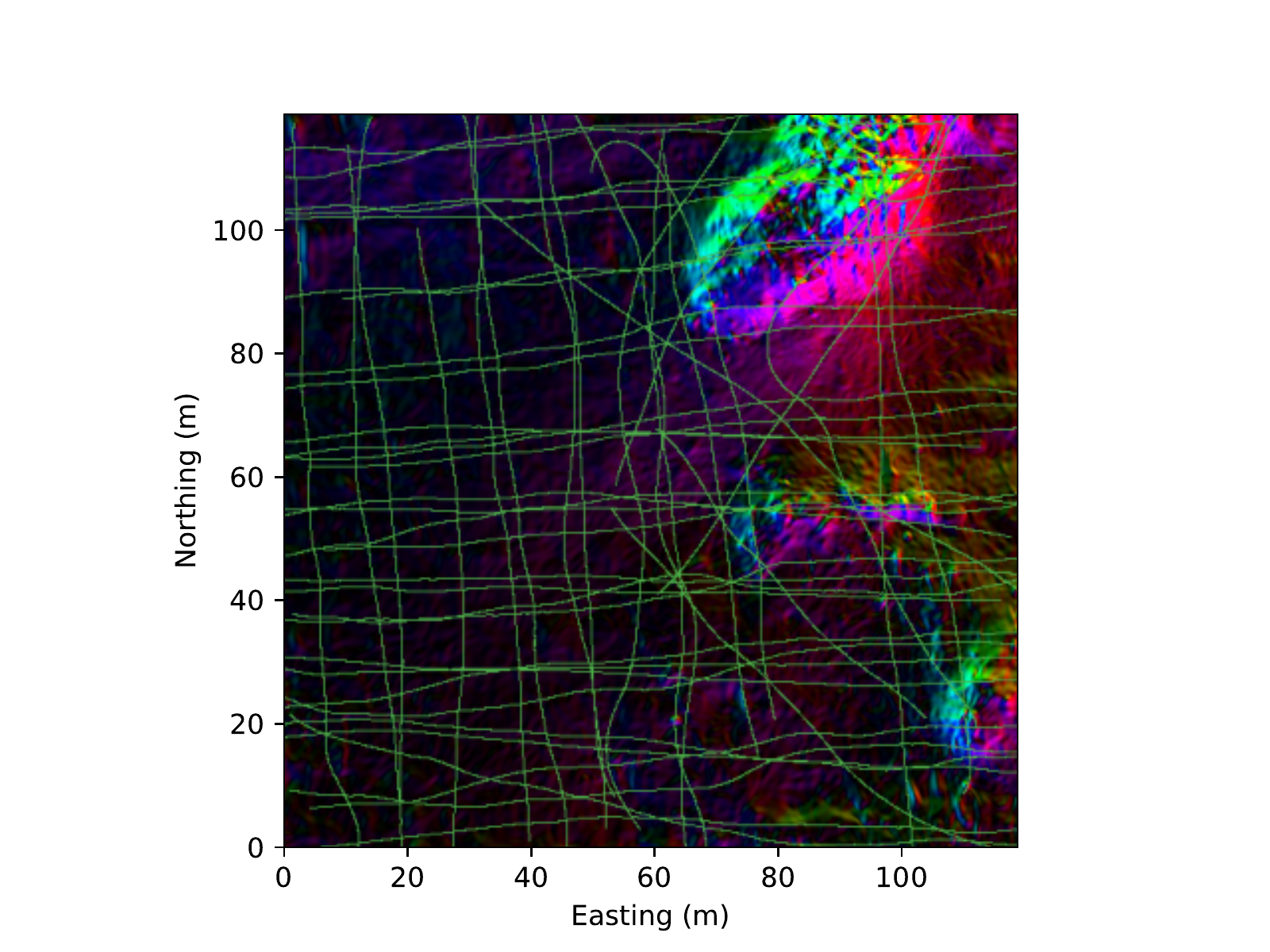}  
  \caption{All altitude readings.}
  \label{fig:densities_all}
\end{subfigure}\\
\begin{subfigure}{.49\textwidth}
  \centering 
  \includegraphics[clip, trim=2.2cm 0.1cm 3.2cm 1.4cm, width=.99\linewidth]{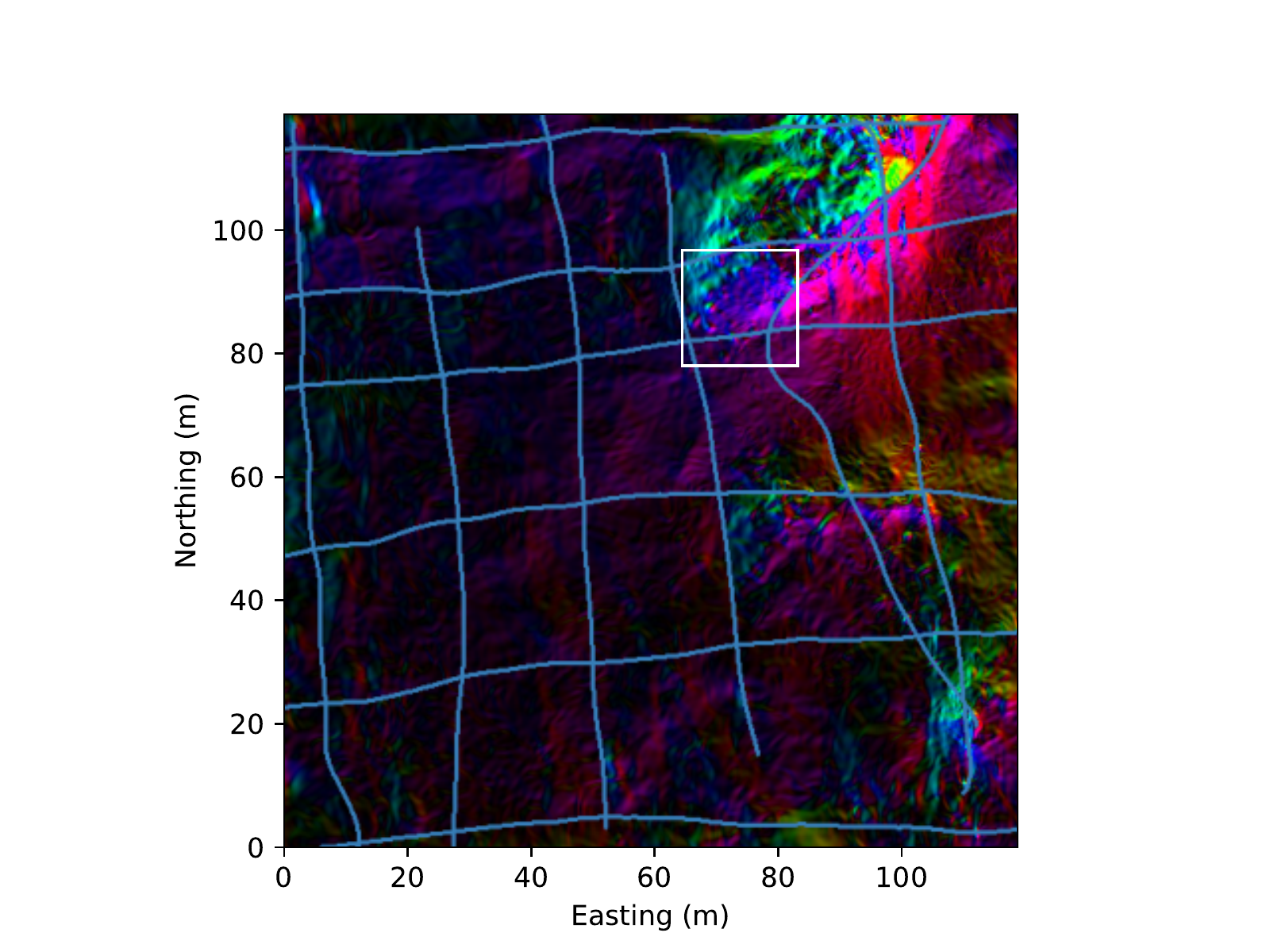}  
  \caption{Sparse altitudes.}
  \label{fig:densities_sparse}
\end{subfigure}
\begin{subfigure}{.49\textwidth}
  \centering
  \includegraphics[clip, trim=2.2cm 0.1cm 3.2cm 1.4cm, width=.99\linewidth]{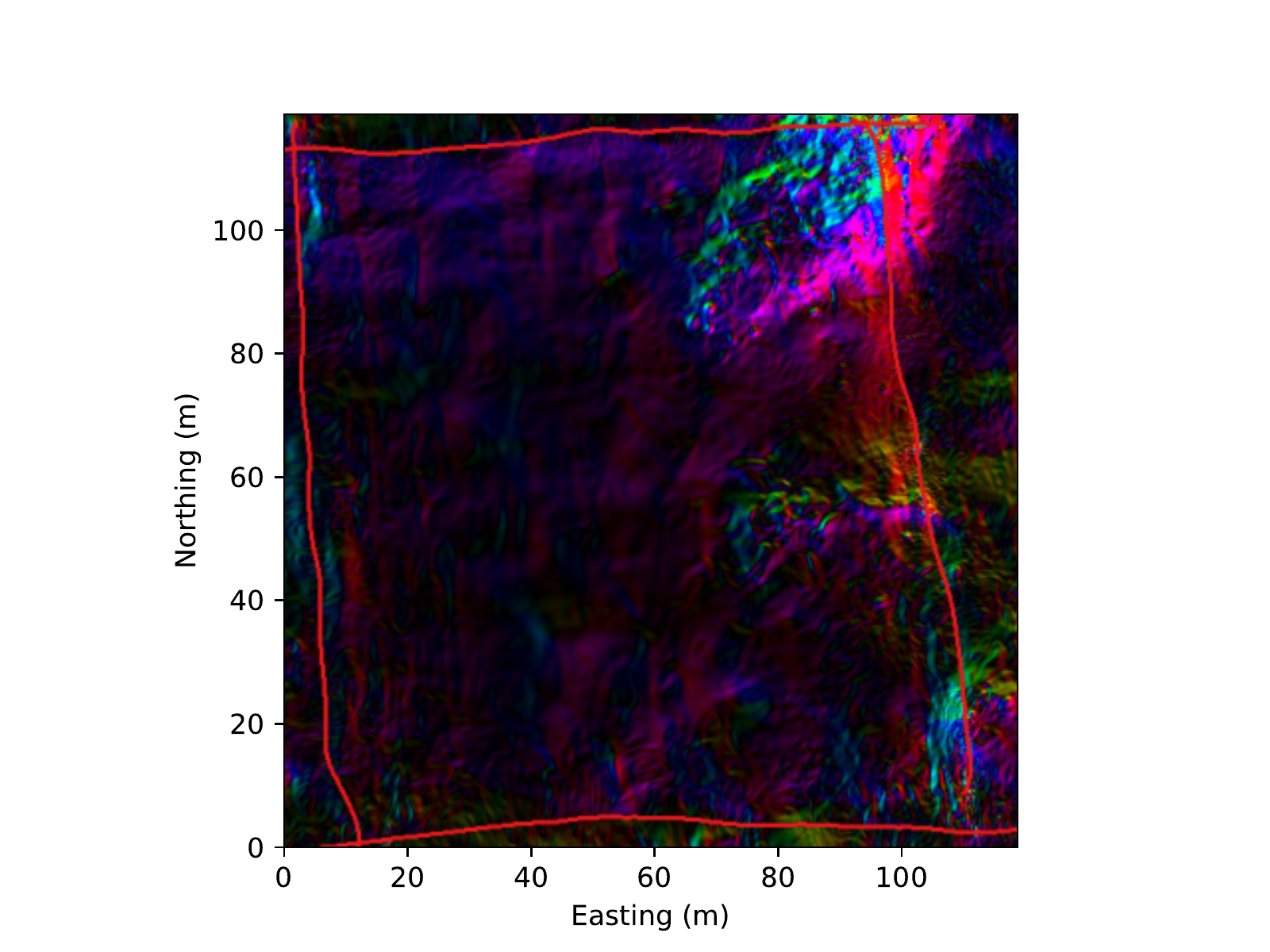}  
  \caption{Border altitudes.}
  \label{fig:densities_border}
\end{subfigure}
\caption{Height map gradient results with different altimeter reading densities (overlaid lines).
         On a large scale, when using all or sparse altitude readings, the method
         manages to reproduce the ground truth bathymetry. With altitudes only at the
         border of the survey area, the absolute values are not accurate but it still
         reproduces the general shape of the hills as well as some rocks. Zoomed in
         example of (a) and (c) (white boxes) can be viewed in Figure \ref{fig:zoomed_densities}.}
\label{fig:densities}
\end{figure*}

\begin{figure*}[ht]
\centering
\begin{subfigure}{.4\linewidth}
  \centering
  \includegraphics[clip, trim=2.0cm 0.1cm 3.2cm 1.4cm, width=.99\linewidth]{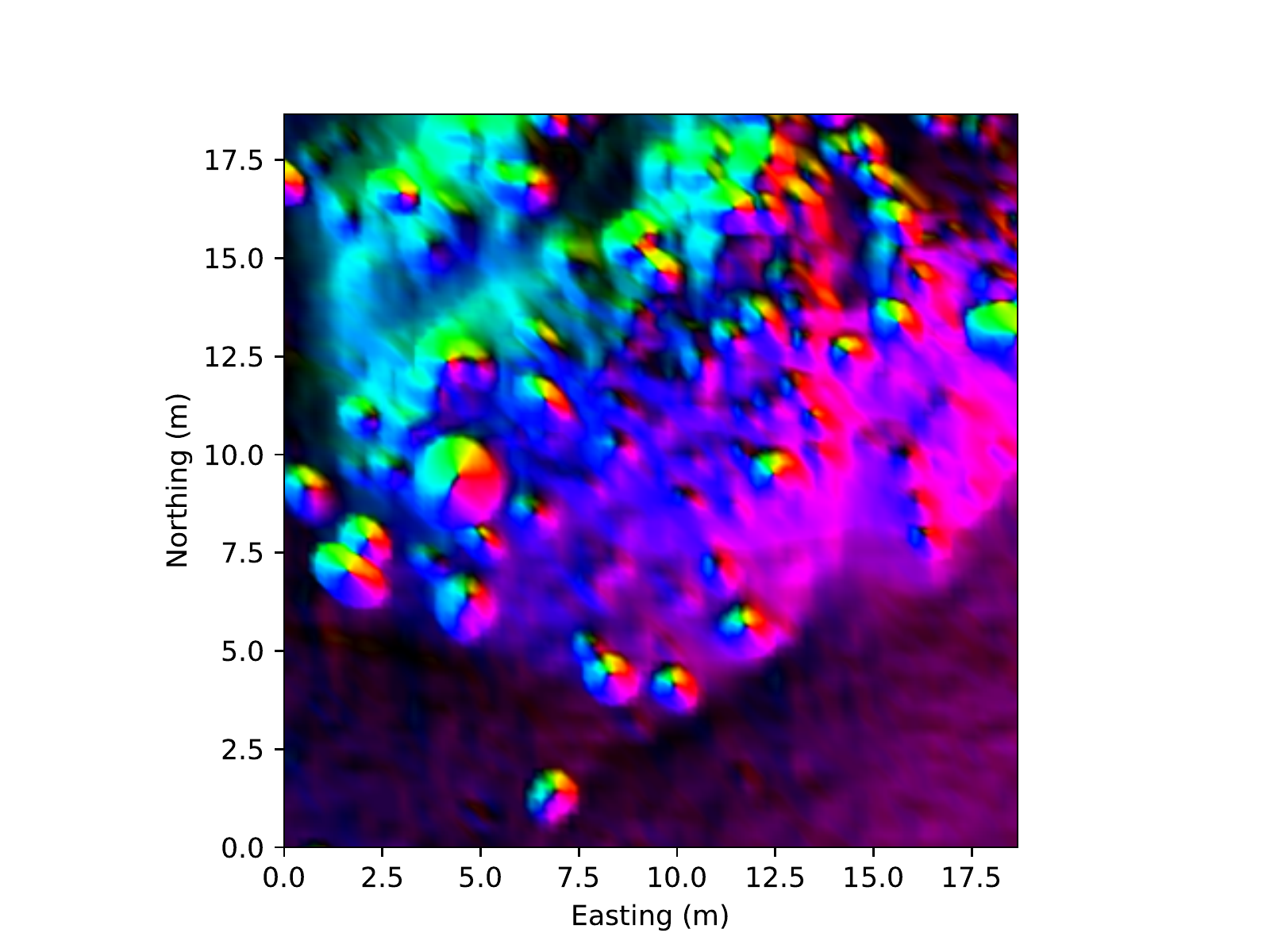}   
  \caption{Ground truth gradients.}
  \label{fig:sub-fourth}
\end{subfigure}
\begin{subfigure}{.4\linewidth}
  \centering
  \includegraphics[clip, trim=2.0cm 0.1cm 3.2cm 1.4cm, width=.99\linewidth]{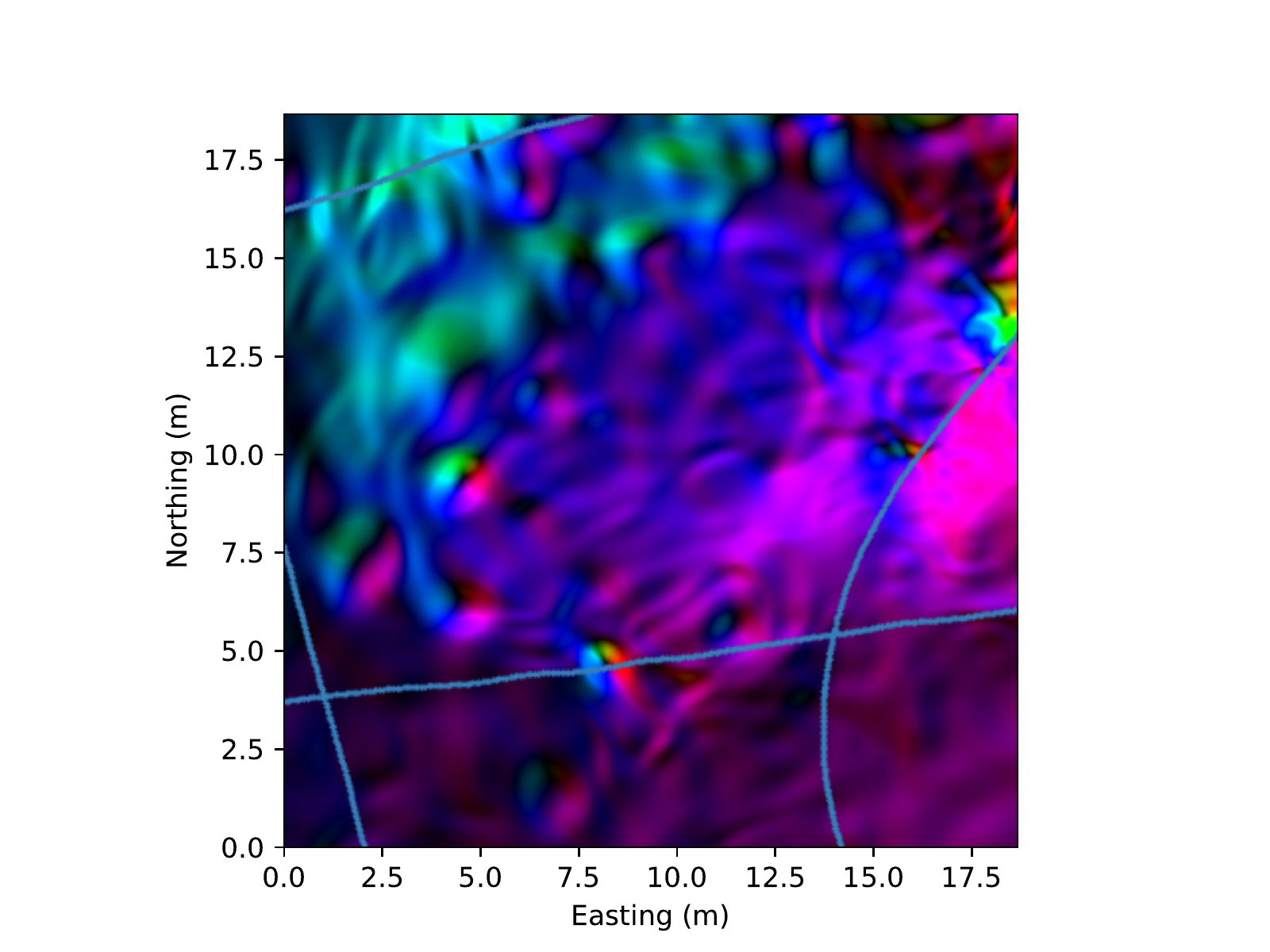} 
  \caption{Sparse altitudes.}
  \label{fig:sub-first}
\end{subfigure}
\caption{Zoomed in examples from Figure \ref{fig:densities}. The model gradients on the
		 right was created from the sparse altimeter readings (blue lines).
		 It manages to reproduce several of the stones not seen in the
		 altimeter data .}
\label{fig:zoomed_densities}
\end{figure*}

The sidescan signal contains roughly 12000 intensities per head.
Our method is able to deal with this high-resolution signal by sampling a
lower number of intensities within each training batch. However, a neural
bathymetry representation with a reasonable number of parameters
is  unable to accommodate this level of precision.
Moreover, the positioning is most likely not accurate to this level of resolution either.
To simplify our analysis, we therefore subsample the signal, keeping only
64 intensities per head in this analysis. Note that the subsampling also
aids in filtering out some of the sidescan noise, thus making the training more efficient.

In order to further accelerate the training,
we produce a naive estimate of the $K$ normalization parameter
from Equation \ref{eq:intensity} for the whole data set.
By assuming a flat seafloor as measured by one altimeter reading below the vehicle,
we can produce Lambertian model intensities for the sidescan pings.
We then fit $K$ to minimize the distance
to the ground truth intensities using a least squares fit on $1\%$ of the pings.
The procedure also ensures that the model works to some extent even without learning
the $A_i$, $R$ and $\Phi$ parameters.

\begin{figure*}[thpb]
	\centering
  	\includegraphics[clip, trim=2.5cm 0cm 3cm 1cm, width=0.99\linewidth]{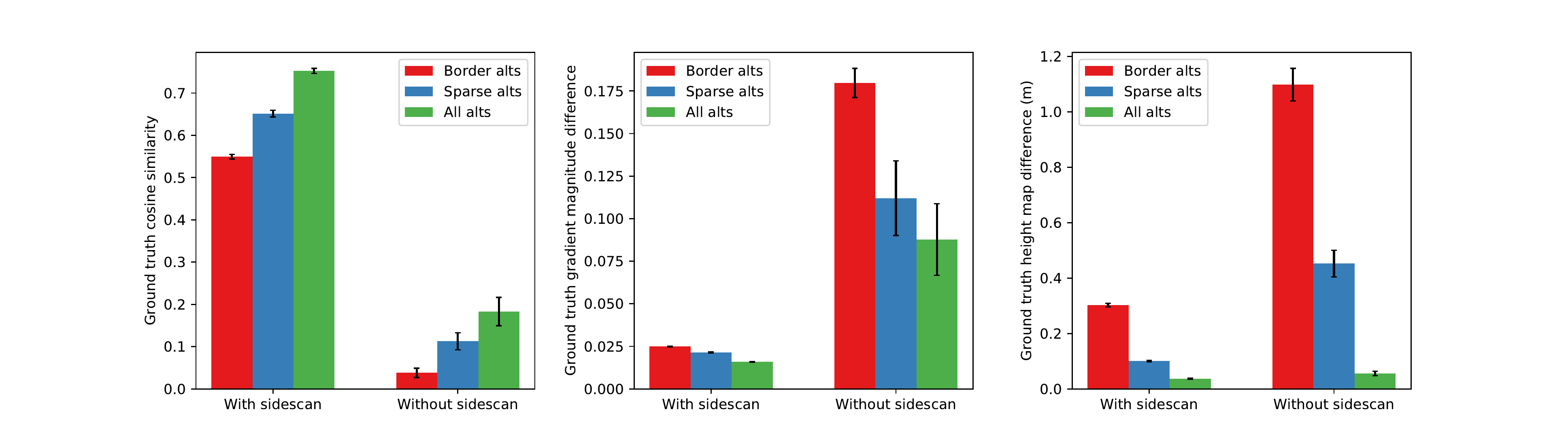}
 	\caption{Comparison of results with and without sidescan as input to the mode,
 			 as well as with different altimeter densities. Sidescan improves the quality
 			 of the reconstruction significantly in all settings and metrics. The difference
 			 is even more pronounced when looking at the gradient quality rather than the absolute height.}
 	\label{fig:density_metric_comparison}
\end{figure*}

Since our method only models reflections from the seafloor, any shadows in the waterfall
images need to be identified and corresponding data points removed for the training.
In our case, we simply remove any intensities below $0.3$ after the initial $K$ correction.
A more sophisticated scheme may also be applied, such as that of Langer \& Hebert\cite{langer1991building}. In our specific data set, there were relatively few shadows,
and our simple method proved effective at detecting them.
Furthermore, we only consider the outer half of the sidescan intensities to avoid
any measurements from the nadir area, which typically occupies the first
third of the ping.

\subsection{Training}
\label{training}

Before feeding into the network, all positional data is scaled by
a factor to make sure that all $x, y$ positions are within the range
$\left[-1, 1\right]$. Since the non-linearity of the network we are using
is the sine function, the normalization ensures that the positions can be
unambiguously represented by the network.

The network used for the experiments is a multi-layer perceptron with 5 hidden
layers, each of width 128.
We train the network for 400 epochs, with an initial learning rate
of $2 \times 10^{-4}$ that is then multiplied by a factor of $0.995$ every epoch.
Each batch contains 400 sidescan pings and 800 altimeter points.
For each ping we sample eight random intensities from each sidescan head,
both port and starboard.

\section{Results}
\label{results}

We start by investigating reasonable values of some of the main
parameters.
These results were generated with the full system, including
albedo, beam-pattern and gain estimation.

\subsection{Parameter search}

\begin{figure}[thpb]
	\centering
  	\includegraphics[width=0.99\linewidth]{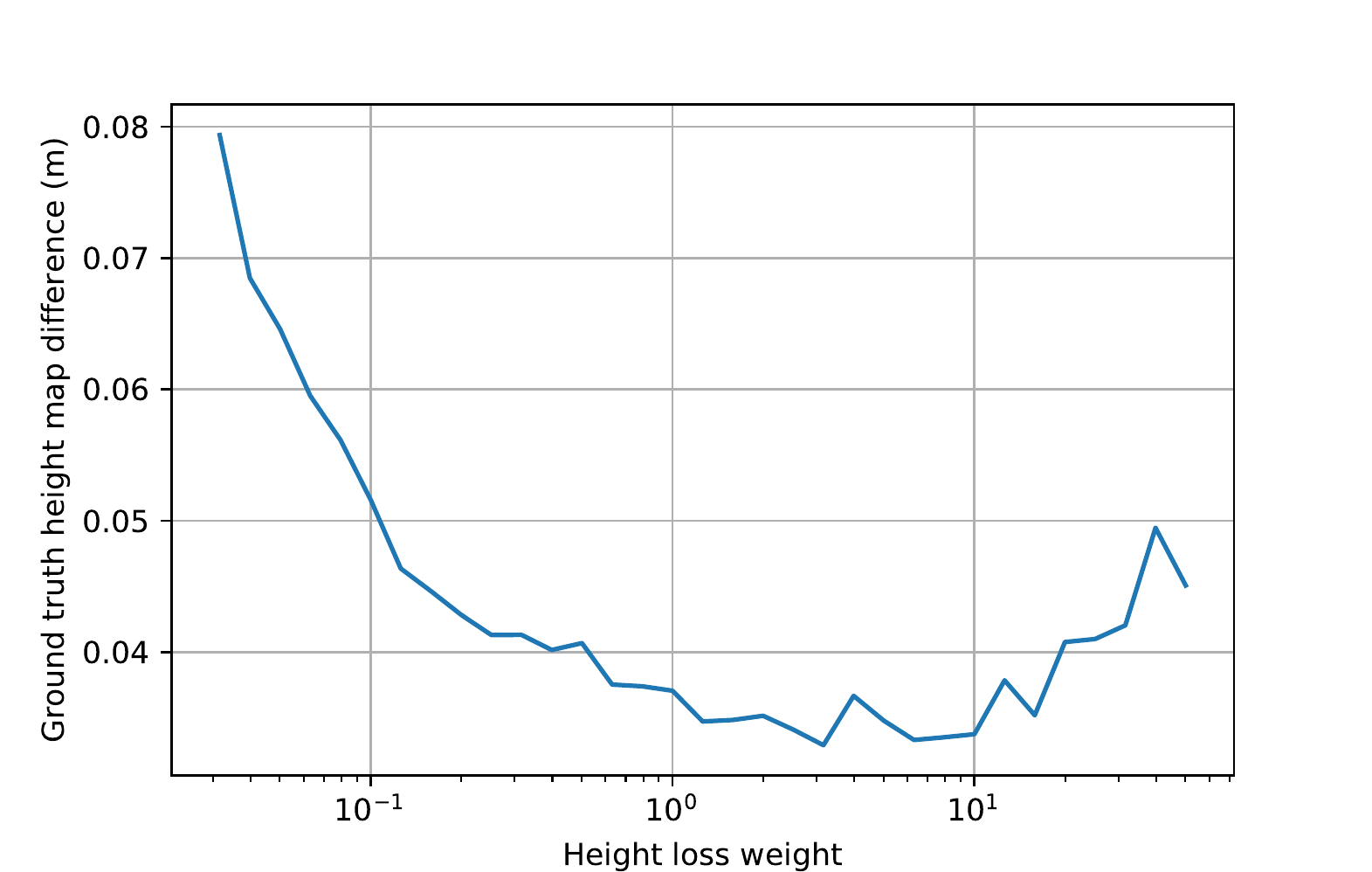}
 	\caption{Mean height map residual as a function of height loss weight $\alpha$.
 	         We observe a minimum around $\alpha=5$.}
 	\label{fig:height_weight_heightmap_diff}
\end{figure}

In Figure \ref{fig:height_weight_heightmap_diff} we can see the results
of varying the height weight parameters $\alpha$. Too much emphasis
on altimeter readings ($\alpha \gg 1$) seems to result in poor bathymetry
estimates. From looking at the produced bathymetries, we conclude that
the network focuses too much on the altimeter readings,
with a somewhat jagged appearance of the seafloor in between.
On the other hand, we also see
that the performance drops dramatically as we approach $\alpha=0$.
In this case, the bathymetry seems to disregard altimeter readings,
instead producing an overly flat seafloor surface. Since no value
seems to significantly outperform $\alpha=1$ that is what we will use in
the remaining experiments.

\begin{figure}[thpb]
	\centering
  	\includegraphics[width=0.99\linewidth]{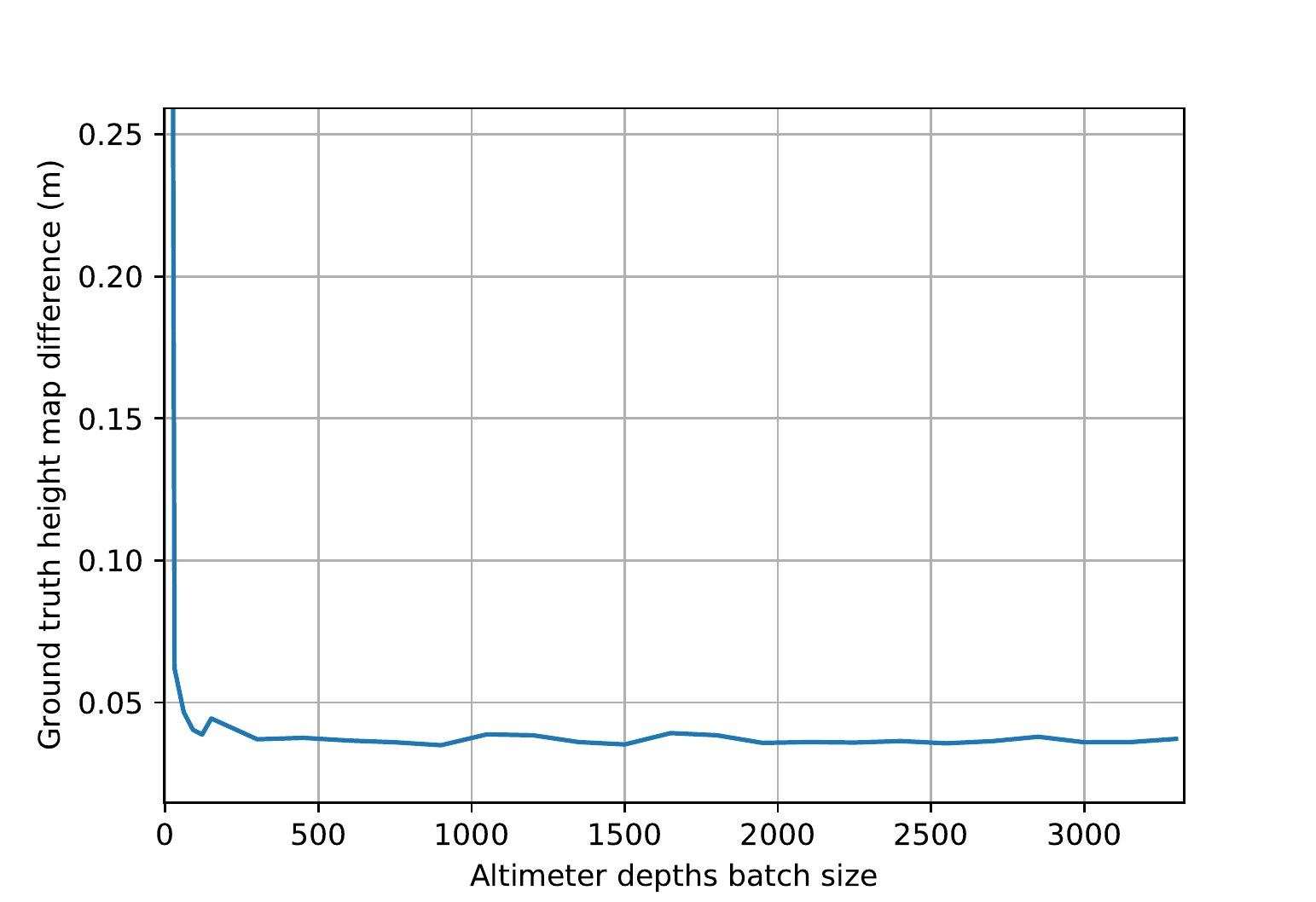}
 	\caption{Mean height map residual as a function of the number of altimeter points
 			 in a batch. The training does not benefit significantly from more than
 			 $\sim 200$ points.}
 	\label{fig:batch_heights_heightmap_diff}
\end{figure}

Next, we investigate the influence of the number of altimeter points
in each batch. The number of sidescan points is kept fixed at 400 since
that is around the maximum number that can be fed through training without
it to slowing down due to the constrained GPU memory bandwidth.
In comparison, it is relatively cheap to add more altimeter points.
However, as can be seen from Figure \ref{fig:batch_heights_heightmap_diff},
there seems to be little gain in supplying more than $\sim 500$ altimeter
points per batch. In the remaining experiments, we use 800 points.

\subsection{No sidescan \& Altimeter sparsity analysis}

We proceed to investigate the benefit of adding sidescan over just using the
sparse altimeter points by themselves.
Figure \ref{fig:density_metric_comparison} illustrates
the improvement, with different altimeter densities.
In general, the pure altimeter height maps still
look reasonable in the sense that the MLP outputs
smooth surfaces in between altimeter points.
However, we observe a significant improvement to including
the sidescan in the optimization.
Interestingly, the improvement is greater
when considering the similarity to the gradients rather than the absolute
height. This is likely due to the sidescan containing information on the gradient,
rather than the height, which may drift when integrating the gradient.

Next, we study the effects of different altimeter densities in detail,
by looking at quantitative as well as qualitative results. We choose three scenarios,
first with altimeter information for \textit{all} sidescan lines.
The resulting gradients are displayed in Figure \ref{fig:densities},
together with the ground truth bathymetry gradients. The general
structure is well approximated, with prominent rocks captured
by the reconstruction. While the large-scale gradients are similar,
the estimated gradient magnitudes are generally somewhat smaller across the
rocks, and many smaller rocks were not identified.

We also test on a \textit{sparse} set containing only 12 lines of altimeter lines, 
with half of the lines being perpendicular to the other half. Again, looking
at Figure \ref{fig:densities} we observe a general similarity to the ground truth
height map. Compared to the results with all lines, there are are somewhat fewer
reconstructed rocks with the sparse lines. However, in the zoomed
in example of Figure \ref{fig:zoomed_densities} we can see that the larger
rocks are reconstructed in an area away from the altimeter lines.
This means those rocks are inferred purely from looking at the sidescan.

In the extreme case of only having  lines at the \textit{border} of the survey area, we can also draw similar conclusions. For example, at the bottom of the
hill of Figure \ref{fig:densities}, we see that the rocks are again reconstructed,
this time far away from any altimeter readings. In general, the height map
seems to degrade as we include fewer altimeter measurements. However, it
maintains a qualitative similarity in between all three densities, with
hills and protrusions appearing in the same places, but with different heights.
In general, our observations agree with the quantitative results
of Figure \ref{fig:density_metric_comparison}.
In summary, the method generates a reasonable estimate even using
very few altimeter points at the borders, that then improves the more altimeter points are added. The best result with all altimeter points exhibit an error
of around $4cm$.

\begin{figure*}[thpb]
	\centering
  	\includegraphics[clip, trim=2.8cm 0cm 3.3cm 1cm, width=0.99\linewidth]{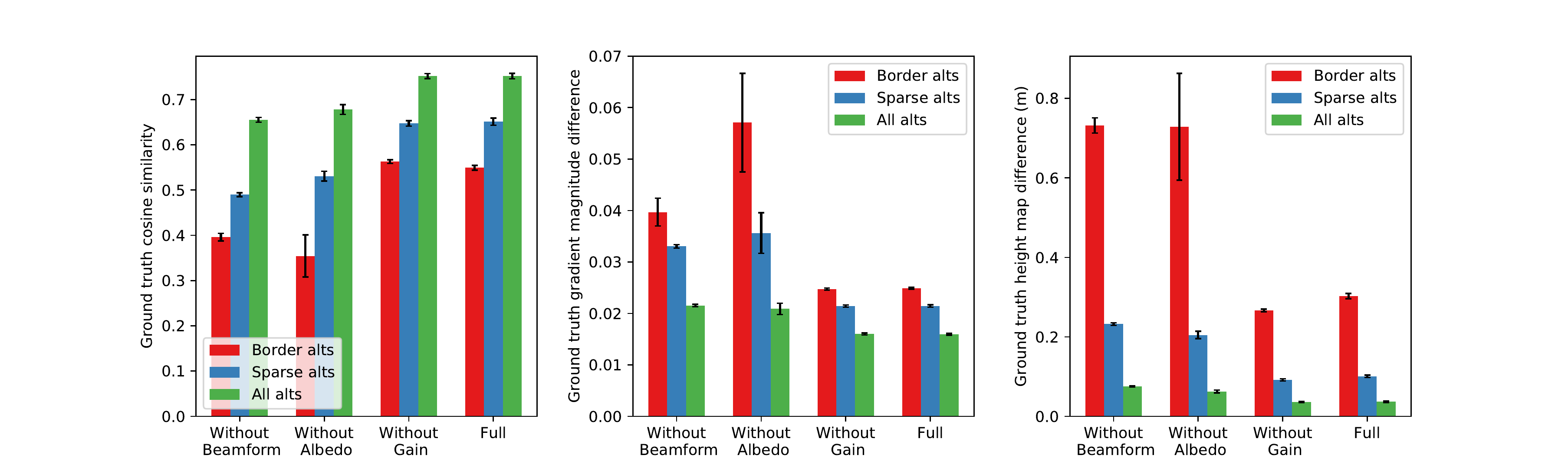}
 	\caption{Ablation study with and without estimating the model parameters,
 			 and with different altimeter reading densities.
 			 Generally, estimating more parameters results in better estimates, with the largest
 			 improvements coming from modeling the beamforms and the seafloor albedo. As may be expected, allowing
 			 for variation in intensity in between the lines does not result in an improvement,
 			 as the gain remained unchanged throughout the survey.}
 	\label{fig:metric_comparison}
\end{figure*}

\subsection{Ablation study}

\begin{figure}[thpb]
	\centering
  	\includegraphics[clip, trim=0cm 0cm 1cm 1cm, width=0.99\linewidth]{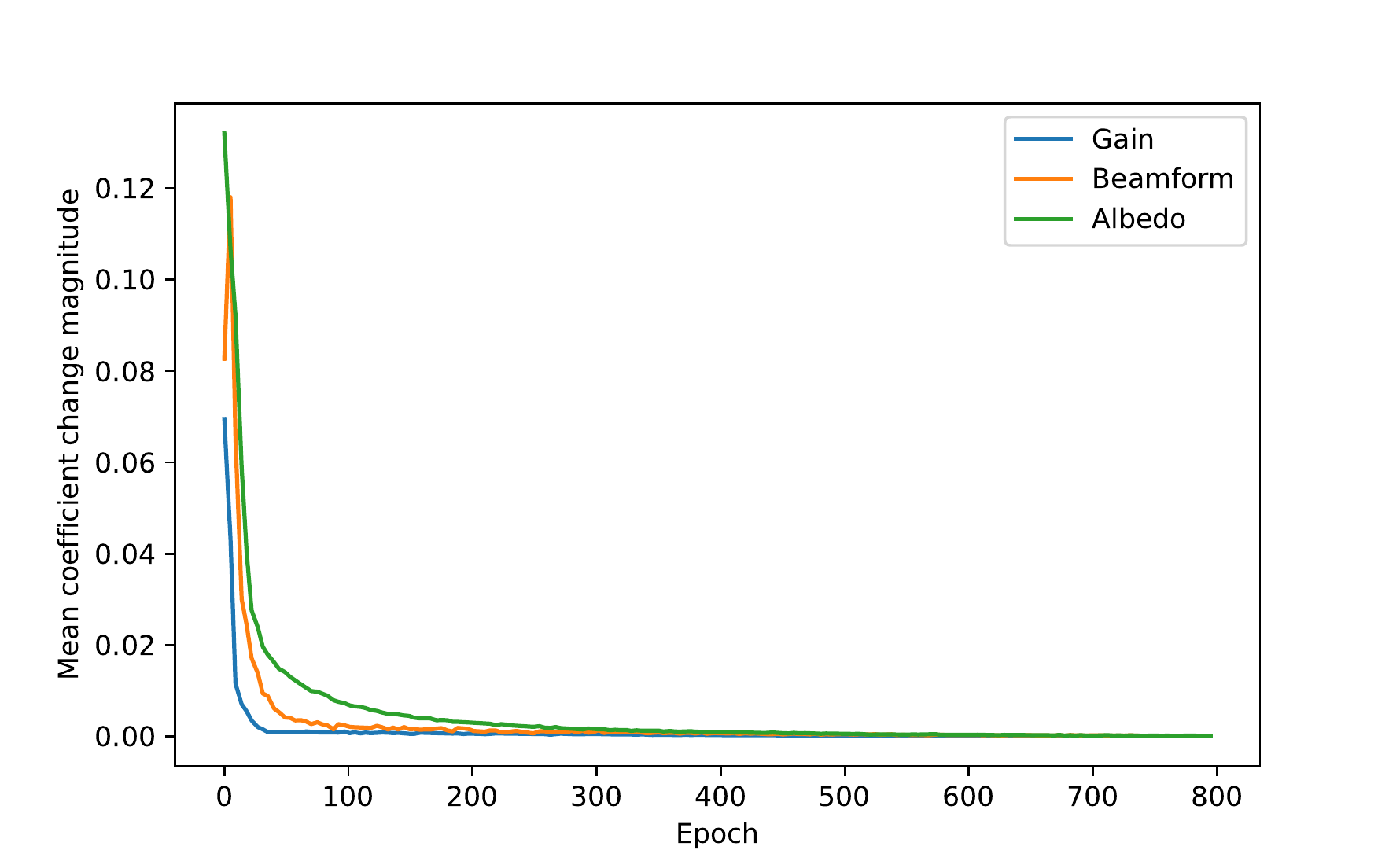}
 	\caption{Mean absolute change of the estimated parameters in between epochs.
 			 All parameters have mostly converged around epoch $\sim 300$.}
 	\label{fig:coeff_change_magnitude}
\end{figure}

In Figure \ref{fig:metric_comparison} we present the results when omitting
estimation of the gain, albedo and beam-pattern parameters.
As the gain did not change throughout the mission, it is only natural that
estimating a gain for each line did not result in an improvement.
Interestingly, it is also not significantly worse than not estimating it,
which suggests that the algorithm may be able to handle a varying gain.
Both albedo and beam-form estimation present a significant improvement in
estimating the height map. This suggests that these two factors play an
important role when sensing our environment. It is also clear that they
are somewhat orthogonal, since estimating both of them at the same time
presents a significant boost. This result also indicates that we had
sufficient coverage of the whole area in our data set, thus avoiding overfit
in some of the parameters, which would result in poor height map estimates.
Looking at Figure \ref{fig:coeff_change_magnitude}, we also conclude that
the estimates converge to stable values when learnt jointly.

\subsection{Performance}

On a computer with an Nvidia GTX 1060, our method is able to
compute $\sim 15 \times 10^4$ isotemporal curve intersections per second,
and backpropagate the corresponding intensity losses, along with the
altimeter losses. With $400$ epochs, 
this takes about $1h$ for the given data set containing $\sim 93000$ pings.

\section{Conclusions}
\label{sec:conclusions}

We presented 
a neural network-based shape-from-shading method that combines sidescan with
sparse altimeter readings to build a dense bathymetry height map.
In experiments, we demonstrated its ability to fuse sidescan data
from a large sidescan survey into a self-consistent bathymetric map.
Notably, it was able to infer the positions
of rocks that were otherwise only visible in the sidescan. When compared to the
ground truth bathymetry from multibeam, the method had an error at the
level of centimeters, which is comparable to the error in the multibeam itself.
Since the depth profile in between altimeter readings
is inferred from the estimated gradient, the estimate sometimes diverges
from the true depth.
Even when the method exhibited quantitative errors, we found that it gives
a valuable qualitative view of the seafloor, clearly indicating the
positions of rocks and ridges.

\section{Future work}

The proposed method is explicitly constructed
to take advantage of the superior resolution of the sidescan
as compared to other sensors. At a fundamental level, it does not
rely on downsampling or discretization. The precision is constrained mainly by the
size of the trainable network, the positioning accuracy, and the aperture of the sensor.
In theory, if the aperture can be modeled within this framework, and if the positioning error could
be mitigated, the model could achieve super-resolution bathymetry.

In practice, the necessary
positioning accuracy can not be achieved at this point, since only the slightest
deviation in angle leads to a significant change in the outer parts of the sidescan ping.
A tantalizing prospect would therefore be to also optimize the sidescan poses together
with the bathymetry. Since the model is fully differentiable, this is possible, but
great care has to be taken for the optimization to still be constrained. In our minds,
an obvious first foray is to optimize the sidescan pose with respect to the vehicle's
navigation frame.

With improved position estimates, it would also be possible to combine sidescan
with a multibeam survey. As discussed, the current method may not reproduce
absolute depths accurately everywhere. If we could use dense multibeam
measurements to fix the height map in more places, that problem may be alleviated.
With sufficiently accurate sidescan positions, our method may be able to enable
resolve smaller details than is currently possible,
while maintaining the same precision as in the coarser multibeam bathymetry.

\section{Acknowledgments}

This work was partially supported by the Wallenberg AI, Autonomous Systems and Software Program (WASP),  by the Stiftelsen  för  Strategisk  Forskning (SSF)  through  the  Swedish  Maritime  Robotics Centre  (SMaRC) (IRC15-0046), and by VINNOVA, project number 2020-04551. Our data set was acquired in collaboration with Marin Mätteknik (MMT) Gothenburg.

\bibliography{paper}
\bibliographystyle{ieeetr}

\begin{IEEEbiography}[{\includegraphics[width=1in,height=1.25in,clip,keepaspectratio]{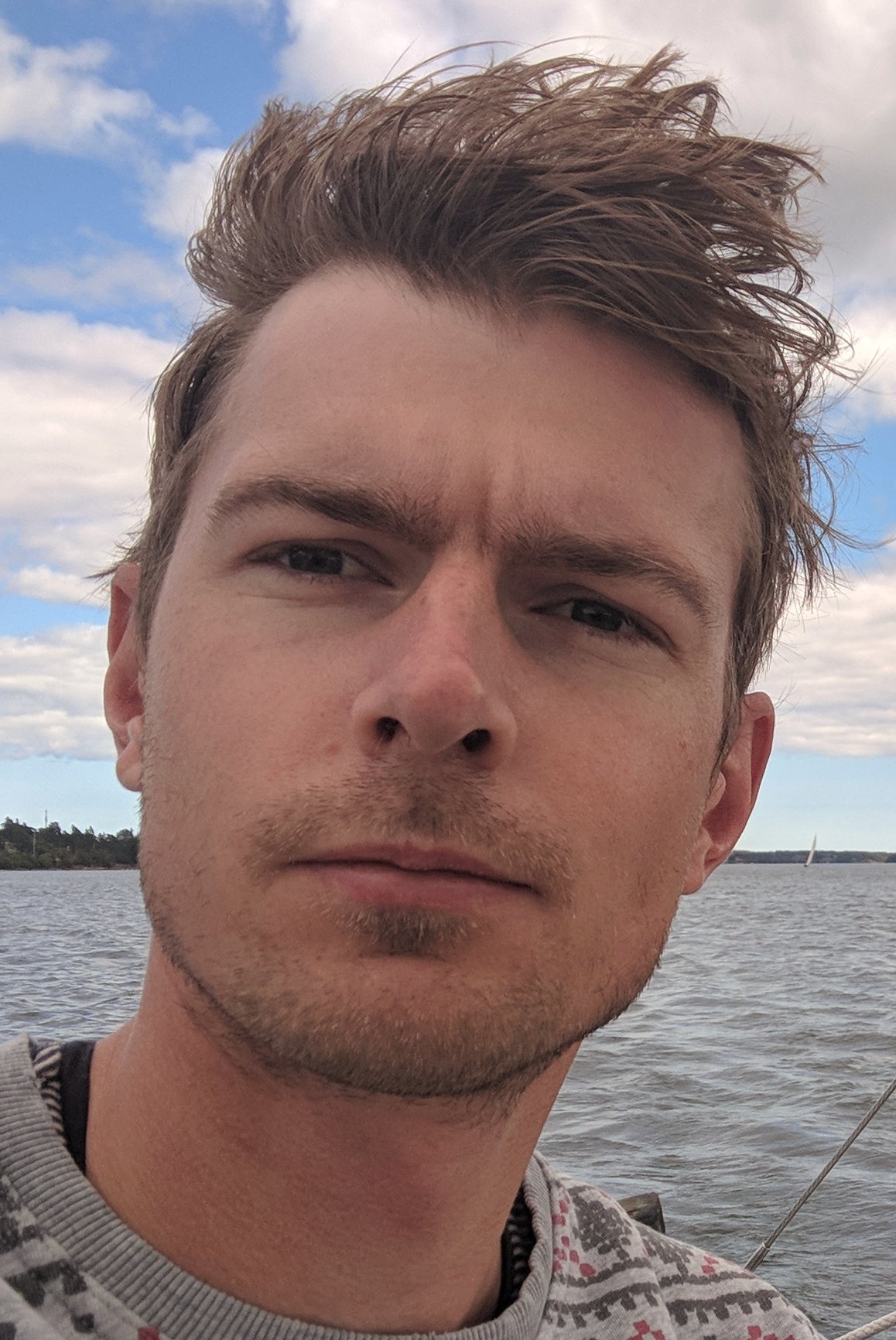}}]{Nils Bore} received the M.Sc. degree in mathematical engineering from the Faculty of Engineering, Lund University, Lund, Sweden, in 2012, and the Ph.D. degree in computer vision and robotics from the Robotics Perception and Learning Lab, Royal Institute of Technology (KTH), Stockholm, Sweden, in 2018.
He is currently a researcher with the Swedish Maritime Robotics (SMaRC) project at KTH. His research interests include robotic sensing and mapping, with a focus on probabilistic reasoning and inference. Most of his recent work has been on applications of specialized neural networks to underwater sonar data. In addition, he is interested in system integration for robust and long-term robotic deployments.
\end{IEEEbiography}

\begin{IEEEbiography}[{\includegraphics[width=1in,height=1.25in,clip,keepaspectratio]{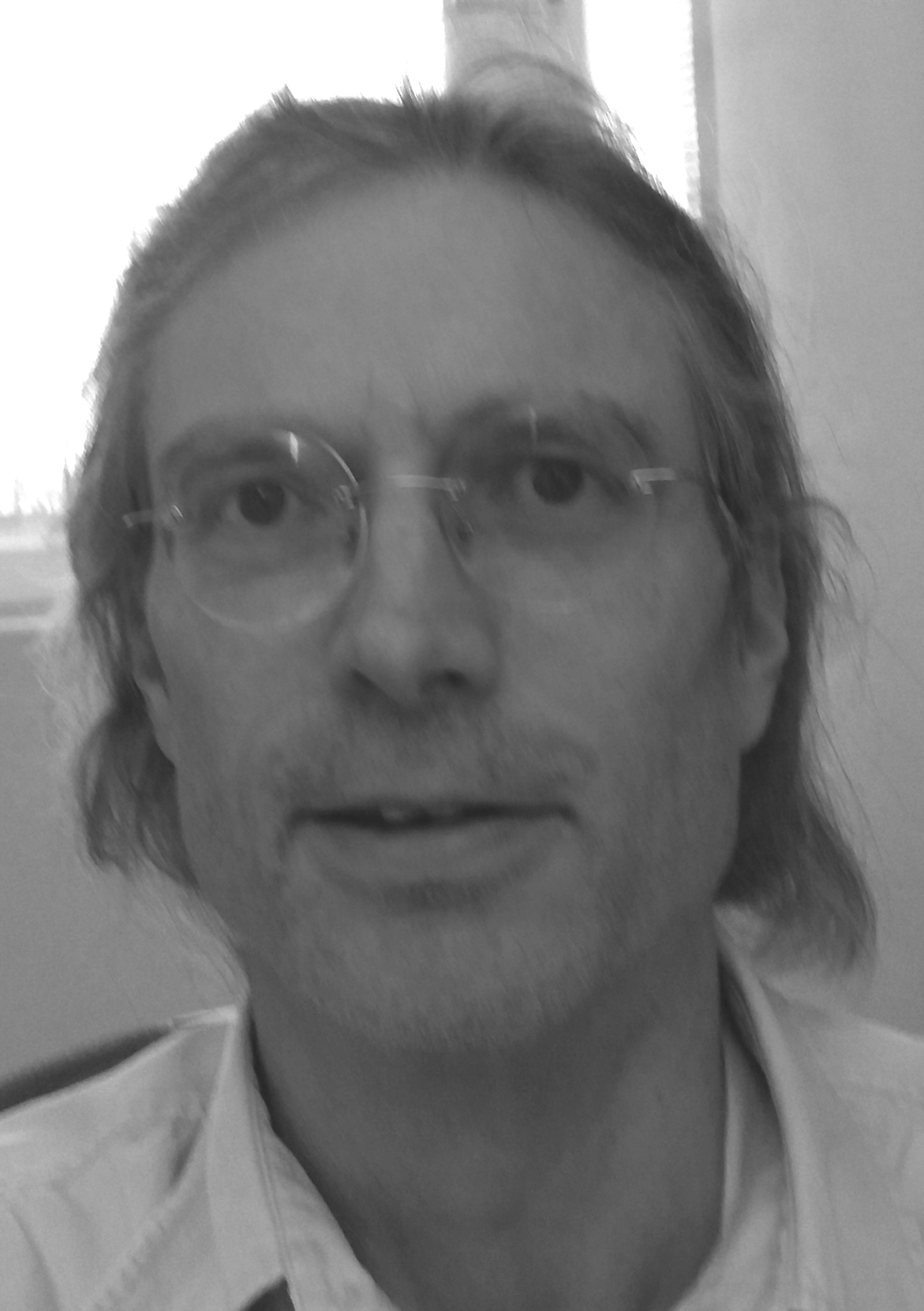}}]{John Folkesson} received the B.A. degree in physics from Queens College, City University of New York,
New York, NY, USA, in 1983, and the M.Sc. degree in computer science, and the Ph.D. degree in robotics
from Royal Institute of Technology (KTH), Stockholm, Sweden, in 2001 and 2006, respectively.
He is currently an Associate Professor of robotics with the Robotics, Perception and Learning Lab, Center for Autonomous Systems, KTH. His research interests include navigation, mapping, perception, and
situation awareness for autonomous  robots.
\end{IEEEbiography}

\end{document}